\def\year{2022}\relax
\documentclass[letterpaper]{article} % DO NOT CHANGE THIS
\usepackage{aaai22}  % DO NOT CHANGE THIS
\usepackage{times}  % DO NOT CHANGE THIS
\usepackage{helvet}  % DO NOT CHANGE THIS
\usepackage{courier}  % DO NOT CHANGE THIS
\usepackage[hyphens]{url}  % DO NOT CHANGE THIS
\usepackage{graphicx} % DO NOT CHANGE THIS
\usepackage{natbib}  % DO NOT CHANGE THIS AND DO NOT ADD ANY OPTIONS TO IT
\usepackage{caption} % DO NOT CHANGE THIS AND DO NOT ADD ANY OPTIONS TO IT
\DeclareCaptionStyle{ruled}{labelfont=normalfont,labelsep=colon,strut=off} % DO NOT CHANGE THIS
\usepackage{algorithm}   % AAD: To remove if problem!
\usepackage{algorithmic}
\usepackage{xspace}
\usepackage[dvipsnames]{xcolor}
\usepackage{soul}
\usepackage[backgroundcolor=white,textsize=tiny]{todonotes}

\usepackage{enumitem}
\setlist[itemize]{leftmargin=*}
\setlist[1]{itemsep=0pt}

\newcommand{\yrv}[1]{{\textcolor{black}{#1}}}

\newcommand{\rebuttal}{\color{black}}
\newcommand{\revisedrebuttal}[1]{\textcolor{black}{#1}}

\newcommand{\tpbox}[1]{\colorbox{white}{#1}} %SpringGreen
\newcommand{\tnbox}[1]{\colorbox{white}{#1}} %YellowGreen
\newcommand{\fpbox}[1]{\colorbox{white}{#1}} %Salmon
\newcommand{\fnbox}[1]{\colorbox{white}{#1}} %YellowOrange
\newcommand{\stext}[1]{\textit{``#1''}\xspace}
\newcommand{\rtext}[1]{\underline{GPT reason}:\textcolor{Black}{\textit{``#1''}}} % changed from Brown color

\usepackage{xcolor}
\usepackage{newfloat}
\usepackage{listings}
\floatstyle{ruled}
\newfloat{listing}{tb}{lst}{}
\floatname{listing}{Listing}
\usepackage{booktabs}
\usepackage{siunitx}

\title{Classifying Conspiratorial Narratives At Scale: False Alarms and Erroneous Connections}
\author {
    Ahmad Diab, %\textsuperscript{\rm 1}
    Rr. Nefriana, %\textsuperscript{\rm 1}
    and Yu-Ru Lin %\textsuperscript{\rm 1}
}
\affiliations {
    School of Computing and Information, University of Pittsburgh\\
    \{ahd23, rrn14, yurulin\}@pitt.edu
}

\begin{document}

\maketitle

\begin{abstract} 
Online discussions frequently involve conspiracy theories, which can contribute to the proliferation of belief in them. However, not all discussions surrounding conspiracy theories promote them, as some are intended to debunk them. Existing research has relied on simple proxies or focused on a constrained set of signals to identify conspiracy theories, which limits our understanding of conspiratorial discussions across different topics and online communities. This work establishes a general scheme for classifying discussions related to conspiracy theories based on authors' perspectives on the conspiracy belief, which can be expressed explicitly through narrative elements, such as the agent, action, or objective, or implicitly through references to known theories, such as chemtrails or the New World Order. We leverage human-labeled ground truth to train a BERT-based model for classifying online CTs, which we then compared to the Generative Pre-trained Transformer machine (GPT) for detecting online conspiratorial content. Despite GPT's known strengths in its expressiveness and contextual understanding, our study revealed significant flaws in its logical reasoning, while also demonstrating comparable strengths from our classifiers. We present the first large-scale classification study using posts from the most active conspiracy-related Reddit forums and find that only one-third of the posts are classified as positive. This research sheds light on the potential applications of large language models in tasks demanding nuanced contextual comprehension.
\end{abstract}

\section{Introduction} \label{sec:intro}

Conspiracy theories,\footnote{\textbf{Disclaimer:} This paper contains analyses and discussions of various conspiracy theories. The inclusion of these theories is solely for the purpose of academic or investigative analysis and should not be interpreted as an endorsement or validation of them.} or CTs have long been the subject of curiosity, interest, and even skepticism \cite{historyCT}. In recent years, the proliferation of CTs has been fueled by the rise of mis- and disinformation on the Internet. While some conspiracy narratives may seem harmless or even entertaining, others can have serious consequences. For instance, conspiracy theories related to COVID-19 \cite{covid_ct}, Pizzagate \cite{pizzaCT, pizzaCT2}, and election fraud \cite{poliCT1, poliCT3} can jeopardize public health, democracy, and public trust. This can result in real-world repercussions such as outbreaks of diseases \cite{vaxCT, preventcovid} and violent insurrections \cite{political_violent}. Therefore, it is crucial to accurately identify conspiratorial content in order to understand the prevailing narratives and mitigate potential consequences.

Conspiracy theories are often convoluted and intricate, involving actors, events, and narratives that imply or explicitly suggest a plot behind the events. They typically lack verifiable details and rely instead on anecdotal evidence, hearsay, or speculation. These characteristics make it extremely difficult to detect conspiratorial narratives in text. In the past, many studies have employed various strategies, such as relying on simple proxies or a small, predetermined set of explicit signals, despite their limitations. For example, a straightforward approach that considers all content from CT-related forums to be conspiratorial \cite{phadke2021characterizing, klein2019pathways, bessi2015science} can produce a large number of false positives, whereas a keyword-driven method that uses predetermined keywords to extract conspiracy theories on a specific topic \cite{kim2023information, hoseini2023globalization} may overlook the false negatives and is frequently limited in scope and generalizability. Other methods, such as pattern matching \cite{kou2017conspiracy, introne2020mapping}, which match textual messages with predefined syntactical elements, can be labor-intensive and may miss narratives that lack exact matching. Recent works have also employed machine learning techniques to automate the classification of CTs, but they are frequently limited to specific topics  \cite{ner_ct, pizzaCT} or lack clear and consistent criteria for identifying conspiratorial content \cite{platt2022toward}. As a result, it is difficult to compare the study results of CT narratives that were extracted using various methods.
\iffalse
In this study, we {\rebuttal{address the shortcoming of previous techniques through developing}} a general classification scheme for {\rebuttal{established and emerging}} conspiracy theories that account for their multifaceted nature, {\rebuttal{relying on our proposed definition in Sec.\ref{sec3.1}: "A conspiracy theory is a set of narratives designed to accuse an agent(s) of committing a specific action(s), which is believed to be working towards a secretive and malevolent objective(s) (secret plot)"}}. Based on the extensive literature on conspiracy theories, our approach considers various aspects of conspiratorial narratives, including (1) the authors' attitudes toward a conspiracy belief {\rebuttal{(e.g., promote, debunk)}}, (2) their explicit articulation of these beliefs through identifiable narrative elements {\rebuttal{(i.e. agent-action-objective)}}, and (3) their implicit alignment with well-known conspiracy theories through references and citations. {\rebuttal{For example, in Sec. \ref{sec7.3}, S2 explicitly references a known CT associated with Dr. Strangelove. Simultaneously, it implicitly alludes to specific elements such as the action (the use of chlorine in water) and the objective (attacking body fluids). The tone of the post indicates the author's supportive and promotional stance toward the theory.}} \fi

\revisedrebuttal{Furthermore, existing approaches that rely on simple proxies or lack theory-grounded criteria can lead to misinterpretation. For example, all discussions about CTs can be mistaken as endorsements of those theories. Our study challenges the appropriateness of previous methods and their limitations in analyzing conspiracy theory prevalence and emergence across a wide range of online forums.}

In this study, we propose a general, topic-independent classification scheme for conspiracy theories that consider their multifaceted nature. Drawing from the extensive research on conspiracy theories, our approach takes into account various aspects of conspiratorial narratives, including the author's perspective towards the conspiracy belief (e.g., promoting or debunking). This can be manifested either (1) explicitly through the use of narrative elements such as the presence of an agent, a particular action, or an objective,\footnote{E.g., ``The government (Agent) is in COVID. They made a larger virus (Action) for population control (Objective).''} or (2) implicitly through referencing and aligning with well-known conspiracy theories.\footnote{E.g., ``Watch out for chem trails (known CT) in the UK. I suspect something is hidden from us.''} We seek to leverage the recent advancements in large language models (LLMs), such as various BERT models and GPT to develop techniques capable of identifying conspiracy narratives within vast online corpora. To this end, we have centered our research on the following key research questions:

%Old content: 
%Drawing from the extensive literature on conspiracy theories, our approach takes into account various aspects of conspiratorial narratives, including (1) the authors' attitudes toward a conspiracy belief (e.g., promoting or debunking the belief), (2) their explicit articulation of these beliefs through narrative elements such as the presence of an agent, a particular action, or an objective, and (3) their implicit alignment with well-known conspiracy theories by referencing well-known theories, such as JFK's assassination or the New World Order.

\begin{itemize}
\item[{\bf RQ1.}] {\it \revisedrebuttal {Can we formulate a \iffalse versatile\fi general, topic-independent, conspiracy theory classification scheme capable of identifying established theories and \iffalse adapting to\fi emerging ones?}}
\item[{\bf RQ2.}] {\it How feasible is it to use large language models (including the BERT family and GPT) to automatically classify online conspiracy theory narratives?}
\item[{\bf RQ3.}] {\it How prevalent are conspiracy theory narratives in conspiracy theory-related forums?}
\end{itemize}

The main contributions of this work include:
\begin{itemize}
\item We establish a general classification scheme that enables the systematic identification of conspiratorial content, taking into account the complex and multifaceted nature of CT narratives. 
\item We develop the first general CT classifier based on the BERT-family of models and demonstrate the effectiveness of incorporating LLMs for detecting online conspiracy narratives. {\revisedrebuttal{Our best model achieved 0.787 AUC.}}
\item We identify the advantages and disadvantages of GPT in comparison to other machine classifiers. While GPT is well-known for its expressiveness and contextual understanding, our results show that our classifiers have comparative strength in terms of classification performance. Furthermore, our analysis of GPT's results provides valuable insights into the challenges posed by generative AI in the context of CT detection. 
\item With the capacity of our best classifier, we present the first large-scale classification study investigating the prevalence of CTs in the most active CT-related Reddit forums. Our analysis reveals that only one-third of the posts are classified as CT narratives, with the remainder of the posts possibly not constituting CTs or not meant to promote CTs. The classification allows us to obtain a more precise picture of the scope and reach of conspiratorial content across online communities. {\rebuttal{Further investigation of the results revealed that posts promoting CT narratives tend to receive more engagement, which suggests that promoters of such theories might have a better chance to further leverage the platform algorithms and promote their content more widely.}}
\end{itemize}

\section{Related Work} \label{sec2}

The analysis of CTs often involves 1) identifying CT text and 2) comprehending its content. Existing works have employed classification techniques or text mining as a part of their analysis. We group current approaches to analysis into four categories: 1) proxy-based CT extraction, 2) keyword-driven approach, 3) analysis with heterogeneous contents, and 4) topic- and cluster-based analysis. 
% \yrl{@NR/AD: you can convert "Papasavva et al. (Papasavva et al. 2021)" to "Papasavva et al. (2021)" by using citet (instead of cite). See the example below. PLz fix the citations throughout the paper to save space.} \ahmad{Done} \yrl{You misunderstood my points. There is a difference between "XX et al.  [CITE]" and "Prior works have done XYZ [CITE]." In the formal case, you use citet to reduce redundant text, but not the latter; otherwise the sentences become grammatically incorrect!! Plz fix them!!} \ana{Done fixing this citation problem for related works, definitions, introduction, ethics, and results, dataset, and machine classification section.}

\paragraph{Proxy-based CT extraction} In previous works, the elicitation of conspiracy messages from social media has often relied on a simple proxy: considering all content from conspiracy-theory-related forums, such as subreddits \cite{phadke2021characterizing, gram_ct,klein2019pathways,engel2022characterizing}, subverses \cite{papasavva2021qoincidence}, or Facebook pages \cite{bessi2015science,zollo2017debunking}, as conspiratorial. As an illustration, \citet{klein2019pathways} investigated the language Reddit users use in a conspiracy forum, r/conspiracy, and the associated social environments. While they implemented a filtration process, it primarily focused on excluding bot accounts and non-active users without involving further CT classification/identification techniques. 

Several studies employing a direct approach have examined banned forums or forums that share names with some banned forums \cite{engel2022characterizing,papasavva2021qoincidence,phadke2021characterizing}. \citet{phadke2021characterizing}, for instance, referenced 17 banned QAnon-related subreddits from various press sources in their investigation of social imaginaries and self-disclosures of dissonance within online conspiracy discussion communities. \citet{engel2022characterizing} analyzed a cohort of users active in 19 QAnon-focused subreddits that were shut down as part of a moderation effort, in their study characterizing the Reddit participation of individuals engaging with QAnon conspiracy theories. \citet{papasavva2021qoincidence} focused on examining QAnon in Voat subverses that shared analogous or identical names with banned subreddits, rather than delving into the banned subreddits themselves.

While this approach has been commonly employed, our empirical evidence, as detailed later, demonstrates that not all posts from these conspiracy-related subreddits can be categorized as conspiratorial (e.g., some users actively debunked conspiratorial narratives; others express skepticism towards a conspiracy theory \cite{tm_ct}). As a result, these false positives could compromise the validity of the subsequent analysis. 

% Take the following post from r/conspiracy\_commons/ as an example.

% Title: “\textit{Solving conspiracies one at a time}”

% Text: “\textit{I've spent the last year and a half pursuing the truth about the events of 2020. Now, I need to get back to studying 9/11.}”

% In this specific conspiracy-related Reddit post, while the author mentions a conspiracy theory topic (the 9/11 conspiracy theories), the narratives in the messages \textit{per se} are not conspiratorial. The author's stance is also ambiguous; they may be actively engaged in debunking. 

% In this regard, the work of Klein, Clutton, and Polito \citet{tm_ct} on modeling r/conspiracy topics has also demonstrated that not every post in this subreddit is conspiratorial. For instance, the authors explained that some of the messages rhetorically capture how some Reddit users converse about conspiracies rather than delving into the particular details of the conspiracies themselves. Additionally, some messages express skepticism towards a conspiracy theory rather than affirming belief in it. This complexity highlights the potential for inadvertently including non-conspiratorial content in a conspiracy dataset, and as a result, these false positives could compromise the validity of the subsequent analysis. 

\paragraph{Keyword-driven approach} Previous research has often relied on a set of predetermined keywords to identify CT instances \cite{kim2023information, phadke2021makes, hoseini2023globalization, pasquetto2022disinformation}. For example, in their investigation of the QAnon conspiracy theory ecosystem on Facebook, \citet{kim2023information} employed 14 core keywords that specifically encapsulated the essential characteristics of QAnon (e.g. ``qanon,'' ``qarmy,'' and ``greatawakening''), alongside 40 extended keywords that encompassed broader aspects of the QAnon narrative and related conspiracy theories (e.g. ``pedogate''). Similarly, \citet{phadke2021makes} utilized the keyword ``conspiracy'' and performed regular expressions to match the string ``conspir'' in subreddits' names and descriptions to find conspiratorial subreddits. By employing Principal Component Analysis and Pointwise Mutual Information, they identified similar subreddits to the already identified conspiratorial subreddits and added them to the list. Then, they categorized all users within the identified conspiratorial subreddits as participants in conspiracy communities. Furthermore,  \citet{pasquetto2022disinformation} harnessed a range of keywords, hashtags, pseudonyms, and symbols frequently used by Italian QAnon influencers on Twitter to indicate their affiliation with the QAnon movement. This approach facilitated the authors in comprehensively studying QAnon influencers' activities by analyzing their tweets and retweets, ultimately aiding them in understanding the intricate disinformation infrastructure of Italian QAnon supporters. However, this approach is often restricted to only one or a few CTs and may not generalize well to different CT topics.

\paragraph{Analysis with heterogeneous contents} Several studies have sought to automate the classification of CT messages using machine-learning methods \cite{ner_ct, pizzaCT, phillips2022hoaxes, pogorelov2021wico, platt2022toward}. However, these works often focus on a limited number of conspiracy topics. For example, when devising a graphical approach to identify conspiracy theories in social media and news automatically, \citet{ner_ct} narrowed their focus to COVID-19 conspiracy theories, while \citet{pizzaCT} limited theirs to Bridgegate and Pizzagate. In a study by \citet{phillips2022hoaxes}, the examination of conspiracy theories was confined to four specific topics (climate change, COVID-19 origin, COVID-19 vaccine, Epstein-Maxwell trial) within their experiment utilizing neural network classifiers for conspiracy, stance, and topic detections. Apart from these studies, there are others that, although not restricting their focus to a specific conspiracy topic, lacked clear criteria for defining what constitutes a conspiratorial message, making it hard to judge the quality of the results \cite{pogorelov2021wico, platt2022toward}. Unlike previous research, we propose a theory-grounded CT classifier applicable for classifying CT texts in various topics.

\paragraph{Topic- and cluster-based analysis} Previous studies have compared the contents between conspiracy and non-conspiracy datasets. In \cite{miani2022interconnectedness}, the authors applied network analysis and text-mining techniques to the LOCO dataset's \cite{miani2021loco} Latent Dirichlet Allocation (LDA) topics. They found that, compared to non-conspiracy texts, conspiracy texts exhibited higher levels of interconnectivity between topics, greater topical diversity, and higher similarity to one another. Similarly, \citet{nerghes2018early}, used topic modeling and semantic network analysis to assess user responses (comments and replies to the comments) to informational (non-CT) and CT videos related to the Zika virus on YouTube. Their research revealed that responses from viewers of informational videos primarily focused on the virus's repercussions, whereas those of conspiracy theory videos also emphasized the parties accountable for the outbreak. \citet{gram_ct} utilized a syntactic rule (a agent-action-target triplet) to extract conspiratorial statements in r/conspiracy. By semantically clustering these triplets into some clusters, they observed ``narrative-motifs'' such as governmental agency–controls–communications.

% \citet{gram_ct} have also worked in the same direction as \citet{nerghes2018early} in examining the nature of conspiratorial discussions in online communities. In their study, Samory and Mitra utilized a syntactic rule, i.e., agent-action-target triplets, and computationally identified them in conspiratorial statements in r/conspiracy. By semantically clustering these triplets into some clusters, they revealed some "narrative-motifs" that generally appear across conspiratorial statements, such as governmental agency–controls–communications, political leader–usurps–power, and country–threatens peace–through military.

\section{Identifying Conspiracy Narratives in Reddit Posts} \label{sec3}

\subsection{Theoretical Definition of Conspiracy Theories} \label{sec3.1}

Existing research offers a variety of perspectives on the precise nature of conspiracy theories, which generally fall into two categories: a) focusing on the constituent elements of conspiracy theories, and b) assessments of the veracity of these theories.

\textbf{Elements of conspiracy theories.} Several scholarly authors \cite{introne2020mapping, ct_df2, mompelat2022loco, wood2015online} have examined the components of conspiracy theories. For example, \citet{introne2020mapping} highlighted six terms contained within a conspiracy theory: 1) events, 2) actors, 3) goal, 4) actions, 5) consequences, and 6) target. In contrast, \citet{ct_df2} argued that the tenet consists of four rather than six basic components: 1) a number of actors joining together, 2) in a secret agreement, 3) to achieve a hidden goal, and 4) which is perceived to be unlawful or malevolent. Aside from these elements, this study added another factor that makes a conspiracy theory dangerous: how the individuals involved in the conspiracy were deviating from their usual behavior. This addition is consistent with the explanation of \citet{mompelat2022loco} on conspiracy belief, stating that causal narratives of an event were not ``random or natural occurrences'' but rather a covert plan carried out by a secret cabal of people or organizations. Meanwhile, in addition to the element of secrecy in how the conspirators carried out their agenda, \citet{wood2015online} also included ``systematic deception'' in their definition. Specifically, they defined a conspiracy theory as ``an allegation regarding the existence of a secret plot between powerful people or organizations to achieve some goal (usually sinister) through systematic deception of the public.''

\textbf{The veracity of the conspiracy theories.} Previous research has also looked into the veracity of conspiracy theories \cite{ct_df3, ct_df1}. A conspiracy theory, according to \citet{ct_df3}, is ``a set of false beliefs in which an omnipresent and omnipotent group of actors are believed to work together in pursuit of malevolent goals.'' According to their definition, theories that turned out to be true, such as the Project MKULTRA and Watergate conspiracies \cite{ct_df1}, are not conspiracy theories. Differently, \citet{ct_df1} acknowledged that conspiracy theories can be true or false, although, in their study on the causes and cures of conspiracy theories, they limited their scope to false conspiracy theories only.

Overall, these prior works shared at least three common elements: agent(s), action(s), and objective(s)/secret plot(s). We further note that conspiracy theories should not be evaluated based on their veracity. Therefore, we define a conspiracy theory as follows:

“\textit{A conspiracy theory is a set of narratives designed to accuse an agent(s) (be they individuals, groups, or organizations) of committing a specific action(s), which is believed to be working towards a secretive and malevolent objective(s) (secret plot).}”

Our definition has elements in common with the work of \citet{gram_ct}, who proposed utilizing an agent-action-target triplet to extract narrative motifs from text content. However, as will be discussed in a later section, our established ground truth is not based solely on the three elements explicitly mentioned in the text, but also on a contextual understanding of the three elements. 

\subsection{Operational Definition of Online Conspiracy Narratives} \label{sec3.2}
While the above theoretical definition serves as a general guideline for identifying conspiracy theories, the identification of conspiratorial content within social media posts has introduced complications. Online conversations are typically informal and highly opinionated \cite{online_opinions}, which can lead to {\rebuttal{long}} posts that do not provide a coherent account of what was discussed {\rebuttal{or address multiple issues deviating from the main narrative of the post, i.e., the most important idea or point that the post is trying to convey}}. The inherent informality and possible loss of context make it more difficult to identify conspiracy theories in social media posts. A post that attempts to debunk a conspiracy theory, for instance, should not be considered a conspiracy theory. Therefore, we propose the following coding instructions {\rebuttal{that help focus on the main narrative in the post by including}} three additional elements:

\textit{``Following the theoretical definition, a social media post that contains a main narrative or claim that (a) represents a known conspiracy theory or (b) suggests a secret plan, along with (c) evidence of agreement or support to some extent for the mentioned conspiracy theory or secret plan.''}

\section{Dataset} \label{sec:data}
Our study focuses on the highly active Reddit communities dedicated to conspiracy theories (CTs). The selection of such communities was guided by \citet{phadke2021makes}, which outline a meticulous manual identification of a core group of popular CT-related subreddits, followed by a systematic search for analogous subreddits based on user engagement and contributions across the platform. 
For each subreddit in the list, we assessed its size by querying the number of posts created within the time frame spanning from 2005 to 2021, utilizing the Pushshift Reddit API. We then used the service's archive\footnote{https://files.pushshift.io/} to download and extract posts published between January 2019 and December 2022 from the 14 largest subreddits. No data was collected after Reddit's new policy change, and we adhere to the platform's data usage guidelines.\footnote{See Sec.~\ref{sec:ethics} for a more detailed discussion of the data access, use, and distribution.}
% For each subreddit in the list, we assessed its size by querying the number of posts created within the time frame spanning from 2005 to 2021, utilizing the Pushshift Reddit API. We then use the service's archive \footnote{https://files.pushshift.io/} to download and extract posts published between January 2019 and December 2022 from the 14 largest subreddits. Unfortunately, Pushshift's API and data dump server are closed due to changes in Reddit's Data API Terms. 
The selected time period captures a sizeable portion of recent online conversations regarding conspiracy theories, allowing for an analysis that reflects evolving trends during this period. Table \ref{table1:dataset} provides a summary of the targeted subreddits and their respective sizes.

The collected data from r/conspiro and r/911truth were significantly lower than their reported sizes. After further investigation, it was determined that r/conspiro was banned at the beginning of 2019, while r/911truth was quarantined, limiting its visibility to users. Subsequently, both of these subreddits were excluded from the study. Furthermore, the data retrieved from r/TopConspiracy, r/conspiracy\_commons, and r/conspiracytheories exceeded the reported figures. This can be attributed to the growing popularity of these forums in recent years, particularly considering that the data collection extended until 2022, whereas sizes are reported up to 2021. Our final dataset consists of 1,122,006 posts from 12 different subreddits.

Some posts have been removed at the time of data collection -- they were either deleted by the users themselves (i.e., {\it self-deleted}) or by any moderation step on the platform (i.e., {\it banned}). The titles of these removed posts may still remain, but their post contents are no longer accessible. Thus we exclude the removed posts in the subsequent analysis. Table \ref{table1:dataset} also lists the number of posts before and after filtering the self-deleted and banned posts.

\begin{table}[t]
\small
\centering
\begin{tabular}{l  r  r r}
\toprule
    \textbf{Subreddit} & \textbf{Size} & \textbf{Full} & \textbf{Clean}\\
\midrule
    conspiracy & 1182794 & 779506 & 201054 \\
    conspiro & 92850 & 142 & 3 \\
    TruthLeaks & 79764 & 60424 & 872 \\
    TopConspiracy & 72389 & 75273 & 311 \\
    conspiracy\_commons & 54437 & 66905 & 12941 \\
    climateskeptics & 51971 & 26091 & 3078 \\
    conspiracytheories & 36436 & 51138 & 11379 \\
    DescentIntoTyranny & 19881 & 11567 & 121 \\
    ConspiracyII & 16756 & 15228 & 1059 \\
    FringeTheory & 15891 & 14749 & 385 \\
    conspiracyundone & 13875 & 9954 & 1420 \\
    C\_S\_T & 13346 & 7126 & 4974 \\
    1984isreality & 12857 & 4045 & 43 \\
    911truth & 12518 & 915 & 199 \\
\midrule
    Total & 1762952 & 1123063 & 237839 \\
\bottomrule
\end{tabular}
\vspace{-0.5em}
\caption{CT-focused subreddits ranked by post activity (2005-2021). ``Full" shows collected posts (2019-2022), while ``Clean" reflects statistics post-filtration.}
\label{table1:dataset}
\end{table}

\section{Establish Ground Truth} \label{sec5}
\subsection{Coding samples} \label{sec5.1}
We took a random sample of posts from three popular subreddits described in Sec.\ref{sec:data}, namely r/conspiracy, r/conspiracy\_commons, and r/conspiracyundone. The sampling occurred after filtering short posts (with less than 30 characters). As shown in Table \ref{table2:annotated}, the final ground-truth data contains 750 coded samples as a result of our human coding process described below.\footnote{Subject to Reddit's terms, the dataset will be made available (see Sec.~\ref{sec:ethics} for details of the data access, use, and distribution).}

\subsection{Human coding process}
In contrast to previous works, which tend to focus on a single or a few topics, we establish a ground truth that encompasses a broad range of CT-related topics. This task requires a team of knowledgeable coders who can comprehend the context of diverse CTs. 
We recruited five coders with prior experience annotating social media texts (e.g., hate speech), including two Ph.D. students, one master's student, and two undergraduates with excellent English proficiency. Among the recruited annotators, two are female. We aimed for disciplinary diversity, with two members from the Computer Science field, two from Information Systems, and one specializing in Digital Narratives. 

%\bigskip
\textbf{Coding instruction.} Each coder underwent training to ensure that they had a comprehensive understanding of the coding guidelines. In addition to the operational definition (Sec. \ref{sec3.2}), we identify four major coding strategies to help coders establish their knowledge of a wide array of CTs and deal with the uncertainty and lack of context in the Reddit posts:

\noindent\textbf{An inventory of known and emerging CTs:} While our definition specifies three elements, many conspiratorial narratives did not explicitly mention all three, instead referring to commonly known CTs (e.g., 5G, NWO, QAnon) or CTs that were becoming popular (e.g., Ukraine biolab, Pizzagate) at the time of posting. Therefore, we compile a list of these CTs{\rebuttal{\footnote{{\rebuttal{The list of CTs, along with the annotation codebook and the dataset are available: \url{https://github.com/picsolab/Conspiratorial-Narratives-At-Scale }.}}}}} and ensure that the coders have a contextual understanding of them. Example: \textit{Be on the lookout for chemtrails in the UK today. I have a theory that something is being hidden from us.} (code: CT).

\noindent\textbf{Rhetorical question vs. genuine inquiry:} Even without explicit language, a rhetorical question (e.g., \textit{Does a lot of conspiracies lead right to Bill Gates? Is he the real leader of the NWO?}) may indicate an author's support for a particular conspiracy theory. We distinguish between rhetorical and genuine questions regarding CTs. Example: \textit{Who's skeptical of the \$1200? What are the odds that they will force you to get the vaccine? Feels like a trap} (code: CT). Example: \textit{Are there any live streams from Afghanistan that are not from a news source? Like people filming right now? Can't find anything on YouTube.} (code: Not CT).

\noindent\textbf{Support/promotion vs. criticism/frustration{\rebuttal{/debunking}}:} CTs related to controversial subjects tend to provoke strong opinions and criticism. However, presenting critical viewpoints and negative sentiments towards controversial subjects in a post does not necessarily qualify it as a CT post, unless the post also expresses endorsement or support for a conspiracy belief. Example: \textit{Oregon has made reading, math, and writing racist which I never thought we could be racist just for breathing! We should all embrace this and bring peace and global health!} (code: Not CT).

\noindent\textbf{Borderline cases:} Deciphering an author's intent presents a great challenge. When a post mentions a CT, but the author's support for the CT is highly ambiguous, the post is deemed non-CT. 
%Example:\textit{ Really interested in pizzagate / child trafficking / andrenachrome conspiracy’s. Wanting to know more truth behind the world's elite; what they really do.} (code: Not CT). 
Example: \textit{Slovakia Covid Testing Video, its cultic as hell and ends with 'papers please'.} (code: Not CT).

{\revisedrebuttal{\iffalse The training encompassed a session with the paper's authors which covered the theoretical and practical definitions of CTs, the labeling strategies previously outlined, and hands-on practice with examples from the dataset labeled by the authors. The five \fi All the annotators underwent a training process that gave them contextual understanding of CTs. This included definitions, emergence, and various examples connecting to recent events. They were also provided with instructions on how to label CTs, and were given at least four hours of hands-on practice with examples from a small dataset labeled by at least two authors on this paper. After the training, the five}} annotators were individually tasked to label each sample as either ``Yes'' (CT) or ``No'' (non-CT) according to our operational definition of conspiracy theory and guidelines. The annotation process can be summarized into three phases, as follows:

\textbf{1) Pilot Phase:} We evaluated the coders' initial agreement after providing them with training on coding guidelines and a series of test cases to ensure they had adequate coding skills. Each coder independently labels 50 samples based on the codebook. The inter-rater agreement between each pair of coders ranged from 0.35 (fair agreement) to 0.80 (substantial agreement) as measured by Cohen's Kappa, and the overall agreement, as measured by Fleiss' Kappa, is 0.54 (moderate agreement). A meeting was held with all coders at the end of this phase to resolve conflicts and disagreements. 

\textbf{2) Consolidation Phase:} While we observed a moderate overall agreement in the previous phase, there was variation in the coding of different individuals. To increase consistency among coders, we divided them into two groups based on their performance in the previous phase. Two rounds of annotations were conducted, each with 100 samples. The coders were instructed to individually label the samples and then to convene within their groups to propose labels that were accepted by the group. Cohen's Kappa values between the two groups for each of the two rounds were 0.65 and 0.74, indicating substantial agreement and improvement over the previous step.

\textbf{3) Conclusion Phase:} Instead of relying on a simple majority vote, the final labels were decided through consensus among the coders. This is to ensure the highest possible coding quality can be reached through the final discussion. All five coders participated in meetings to resolve disagreements. During the meeting, coders defended their annotations and engaged in a discussion to reach a final consensus on the labeling of each sample. We compared the original annotations of each coder with the final agreement to assess the coders' reliability. The remaining 500 samples were coded by two of the coders who demonstrated the highest reliability in the earlier phase, and the final labels were determined by consensus.

\begin{table}[t]
\small
\centering
\begin{tabular}{l m{3.5em}  m{3.5em} r}
\toprule
    \textbf{Subreddit} & \textbf{CT Count} & \textbf{non-CT Count} & \textbf{Total} \\
\midrule
    conspiracy & 100 & 204 & 304 \\
    conspiracy\_commons & 90 & 208 & 298 \\
    conspiracyundone & 58 & 90 & 148 \\
\midrule
    Total & 248 (33\%) & 502 (67\%) & \textbf{750} \\
\bottomrule
\end{tabular}
\vspace{-0.5em}
\caption{Ground-truth labels from human coding.}
\label{table2:annotated}
\end{table}

\section{Machine Classification} \label{sec6}
Based on the human-annotated ground-truth samples, we develop machine classifiers to automatically classify a given post as CT or not. We extensively explore various approaches, including traditional machine-learning methods (ML), deep-learning models that incorporate large language models (LLMs), and a state-of-the-art generative model, the Generative Pre-trained Transformer Machine (GPT). {\revisedrebuttal{\iffalse For both ML and LLMs experiments, we used a shuffled and stratified 5-fold cross-validation technique with 80\%:20\% train:test splits.\fi For both ML and LLMs experiments, we report the 5-fold cross-validation results with a 80:20 training/testing split.}}

\subsection{Deep Learning Models (DLs)} \label{sec6.1}
Several pre-trained LLMs have demonstrated outstanding performance in a variety of NLP tasks. We leverage the capabilities of these LLMs, namely BERT (Bidirectional Encoder Representations from Transformers) \cite{bert}, ALBERT (A Lite BERT) \cite{albert}, DistilBERT (Distilled version of BERT) \cite{distilbert}, DeBERTa (Decoding-enhanced BERT with disentangled attention) \cite{deberta}, RoBERTa (Robustly optimized BERT approach) \cite{roberta}, and T5 (Text-to-Text Transfer Transformer) \cite{T5}, to develop deep-learning models by adding a sequence classification head on top with cross-entropy loss. We build our models utilizing Hugging Face architectures and optimize and fine-tune the classification parameters based on the labeled samples.\footnote{We used the Adam optimizer and the optimal hyperparameters (batch-size:32, lr=1e-5) with 10 epochs in the experiments. All models were trained on a single GeForce GTX TITAN X 12GB GPU. The total training time took approximately 4 hours.}

\subsection{Generative models (GPT)}
% We evaluate the performance of one of the cutting-edge generative models, GPT, in our CT classification task. We utilize OpenAI's APIs with carefully designed prompts\footnote{Specifically, in this study, we used the model \texttt{gpt-3.5-turbo}, accessed in September, 2023. However, our pilot study indicates that the results produced by GPT-3.5 and GPT-4 do not have significant differences.}. Recent research has demonstrated that GPT's performance can be enhanced by providing a few labeled examples (i.e., a few-shot setting) \cite{fewshot_learners} and by utilizing various prompting strategies \cite{CoT}. Following recent reseach \cite{all of them}, we explore with prompts with different strategies and in-context examples fewshot. 
In this study, we utilize OpenAI's APIs\footnote{Specifically, in this study, we used the model \texttt{gpt-3.5-turbo}, accessed in September, 2023. However, our pilot study indicates that the results produced by GPT-3.5 and GPT-4 do not have significant differences.} to evaluate GPT's performance in classifying online CT content. Recent studies highlighted the important role of prompt design when leveraging GPT \cite{fewshot_learners, CoT}. For instance, \citet{choosing_examples} suggested that augmenting prompts with semantically similar examples can lead to performance improvements, while \citet{GPTstrategies} discussed designing strategies that include manual \cite{CoT} or template-defined \cite{zeroshot_reasoner} steps for clarifying the procedure to attain favorable results. In our study, we design three prompting strategies, in zero- and few-shot settings:{\rebuttal{\footnote{{\rebuttal{Examples used in the few-shot setting are available: https://github.com/picsolab/Conspiratorial-Narratives-At-Scale.}}}}}

\textbf{1) Simple:} Asks the model to decide the text's label using the prompt:
\textit{Decide whether the following text describes a conspiracy theory or not (yes/no). ``[post text]''}.

\textbf{2) Justification:} Asks the model to judge the text's label and provide a justification for the label. Prompt:   
\textit{Decide whether the following text describes a conspiracy theory or not (yes/no). Justify your answer. ``[post text]''}.

\textbf{3) Step-By-Step (SBS):} {\rebuttal{Guides the model in determining the label of the text by providing step-by-step instructions designed to replicate the decision-making process instructed to human coders during annotation, with the final step asking whether the post is CT or not}}. Prompt: 
\textit{Decide whether the following text describes a conspiracy theory or not (yes/no). First, extract the narrative or claim from the text. Second, decide if the claim is a known conspiracy theory or suggests a hidden plan. Third, decide if the text agrees with or supports the conspiracy theory or plan. Fourth, answer the question (yes/no). ``[post text]''}.

%previous first sentence: To implement the few-shot settings, $n$ examples with their respective ground-truth labels are provided to the model.
To assess GPT's performance in few-shot settings, prompts are augmented with $n$ examples, each paired with its respective ground-truth label. In each few-shot setting, a set of examples, both from CT and non-CT labeled samples, are selected based on their similarities to the text under consideration. We experiment with $n=0,1,3,5$ pairs, where $n=0$ represents a zero-shot setting, and the few-shot settings provide the model with 2, 6, and 10 examples, ordered randomly to mitigate the influence of example arrangement \cite{choosing_examples}. We compute pairwise cosine similarity based on the text embeddings generated by the best-performing LLM models (Sec.\ref{sec6.1}) in our experiments (i.e., RoBERTa, as reported in Sec.\ref{sec7.1}). The most similar examples with respect to a given text were then extracted from the labeled data's positive and negative samples. Throughout the experiments, the parameter max. token length is set as 1500 and the temperature is 0, which is a recommended value for classification tasks that effectively constrain randomness. Each experiment was repeated 10 times to provide robust results that accounted for GPT's randomness.

\subsection{Traditional Machine Learning Models (MLs)}
We test commonly used supervised machine learning algorithms, including Logistic Regression (LR), Support Vector Machine (SVM), Decision Tree (DT), Random Forest (RF), K-Nearest Neighbors (KNN), and eXtreme Gradient Boosting (XGB). The implementation of these models was executed using Python's scikit-learn library. For XGB, we utilized the XGBoost package and conducted hyperparameter optimization through grid search. All these ML models are trained with various text embeddings, including all the LLMs described in Sec.\ref{sec6.1}.

\section{Results} \label{sec7}
\subsection{Detection Performance} \label{sec7.1}
% \yrl{@AD: use Capitalized Precision, Recall, F1, and AUC consistently in the paper. Remove the word "score" as it's not necessary when mentioning these metrics.}\ahmad{Done}
Table \ref{table3:results1} presents the Precision, Recall, F1, and AUC of the Machine Learning classifications with the best text embeddings, and Deep Learning models. SVM and XGB consistently demonstrate strong performance across all text embeddings. LR exhibits high Recall, but relatively lower Precision, resulting in lower F1. Notably, RF assigns all samples as positive, thereby achieving perfect Recall but compromised Precision.
In terms of text embeddings, ALBERT, the most compact model among those tested, exhibits relatively lower performance. In contrast, DistilBERT outperforms both BERT and DeBERTa, despite its smaller size owing to distillation. RoBERTa and T5 consistently deliver superior performance across all evaluated metrics.

Among the Deep Learning models, RoBERTa emerges as the top performer, recording the highest F1, AUC, and Precision. Following closely are DeBERTa, BERT, and DistilBERT, which produce comparable results. ALBERT, although exhibiting lower performance, outperforms T5, which surprisingly ranks as the least effective model. We posit that T5's underperformance may be attributed to the need for a larger training dataset and further exploration of hyperparameter optimization.

\begin{table}[t]
\small
\centering
\begin{tabular}{l r r l l}
\toprule
    \textbf{Model} & \textbf{Precision} & \textbf{Recall} & \textbf{F1} & \textbf{AUC} \\
\midrule
\multicolumn{5}{c}{ML Models}\\
    DT+RoBERTa  & 57.6\% & 52.1\% & 0.497 & 0.605 \\
    RF+RoBERTa  & 35\% & 100\% & 0.519 & 0.5 \\
    LR+RoBERTa  & 42.4\% & 79\% & 0.553 & 0.609 \\
    KNN+T5  & 54\% & 69.6\% & 0.608 & 0.689 \\
    SVM+T5  & 79.1\% & 54.7\% & 0.647 & 0.735 \\
    XGB+T5  & 72.5\% & 66.2\% & 0.692 & 0.763 \\
\midrule
\multicolumn{5}{c}{DL Models}\\
    T5  & 33.8\% & 96.5\% & 0.5 & 0.5 \\
    ALBERT  & 58.9\% & 54.3\% & 0.565 & 0.673 \\
    DistilBERT  & 63.9\% & 69.1\% & 0.664 & 0.745 \\
    BERT  & 66.7\% & 70\% & 0.673 & 0.752 \\    
    DeBERTa & 66.1\% & 71.5\% & 0.687\ & 0.762 \\
    RoBERTa  & 70\% & 73.8\% & \textbf{0.714} & \textbf{0.787} \\
\bottomrule
\end{tabular}
\vspace{-0.5em}
\caption{Performance metrics of the ML and DL classification models. Best performances in bold. 
% \yrl{@AD: Use the correct abbreviations DLs.}\ahmad{Done} \yrl{@AD: move AUC to the last column after F1 for both Table 3 \& 4. (I thought I already gave this comment before?)}\ahmad{Done.}
% \yrl{@AD: It looks like DeBERTa has higher F1. It's very unlikely because it's AUC is very low. Could you check?} \ahmad{fixed}
}
\label{table3:results1}
\end{table}

\subsection{Analysis of BERT-based Models' Results} 
{\rebuttal{To assess the precision of BERT-based models, particularly the top-performing RoBERTa (with an AUC of 0.787), we conduct a qualitative error analysis. While RoBERTa exhibited superior performance compared to other models, it inevitably encountered challenges in accurately labeling certain samples. This section presents instances of both false positives and false negatives. Note that in our experiment, GPT misclassified all the following examples.

\paragraph{False Positive} We observed that the RoBERTa model has a tendency to label a given sample as positive when it contains a known CT, irrespective of the post author's intended sentiment.
\begin{itemize}
\item[{\bf I1}] \tpbox{FP} (Label: No; RoBERTa: Yes) \stext{This sub has become cancer. It is a dumping ground for Qanaon retards and pro Biden shills. Biden and Trump are equally evil. Trump is the vax daddy. Biden diddles kids if you deny either youre blissfully ignorant.}
%\item[{\bf I2}] \tnbox{FP} (Label: No; RoBERTa: Yes) \stext{According to David Icke I have reptilian illuminati heritage because of my Plantagenet ancestry. What special powers should I have? What special powers should I have? Is this why street lights seem to turn off when I approach them?}
\end{itemize}

\paragraph{False Negative} We observed that the RoBERTa model has limitations in identifying new and emerging narratives. For example, the author in \textit{I2} below constructs a novel narrative on gene alteration implying it is a made-up plan.
\begin{itemize}
\item[{\bf I2}] \tpbox{FN} (Label: Yes; RoBERTa: No) \stext{I cant believe Im watching the last battle of current humanity. If vaccines are gene therapy or some gene altering something, well be the last human beings of old world. New generations all will have different genes. Mutated, damaged, who knows?And if microchipdigital currency replaces current monetary system, its end of old system too.}
%\item[{\bf I4}] \tnbox{FN} (Label: Yes; RoBERTa: No) \stext{Americans have been so groomed they dont even care when bureaucrats lie to them. Cop beats a man.Police release official statement denying it Video shows it happenedRinserepeat But BLM is the problem, not the cops gt;the officer took the suspect down onto a patch of grass and handcuffed him without further incident. At no time did the officer strike the suspect.}
\end{itemize}
}}

\subsection{Analysis of GPT's Results} \label{sec7.3}
% \yrl{@AD: it's inaccurate to use ChatGPT; use GPT instead. (I've noticed that this comment has been made before.  It would be appreciated if you could make an effort to avoid making the same mistakes again.)} \ahmad{Changed. and Noted.}
Table \ref{table4:GPTresults} provides an overview of GPT’s performance across various experimental settings, each experiment was repeated 10 times to alleviate GPT’s randomness. The Standard Deviation across all metrics ranges between 0.002 and 0.042. Notably, the ``Simple'' prompt setting slightly outperforms both the ``Justification'' and ``Step-By-Step'' settings. However, the absence of reasoning in this setting casts doubt on the final verdict, particularly considering the alternation of labels when justification was requested. 

%However, it is worth emphasizing that the binary labeling nature of the task may restrict interpretability, particularly considering evidence suggesting that GPT changed its judgment on some samples when asked to provide reasoning.

\begin{table}[t]
\small
\centering
\begin{tabular}{l l l l l}
\toprule
    \textbf{Setting} & \textbf{Precision} & \textbf{Recall} & \textbf{F1} & \textbf{AUC} \\
\midrule
    Simple &  &  &  &  \\
    \begin{small}\space\space0-shot\end{small} & \textbf{72.50}\% & 72.60\% & \textbf{0.726} & \textbf{0.795} \\
    \begin{small}\space\space1-shot\end{small} & 65.60\% & 76.90\% & 0.708 & 0.785 \\
    \begin{small}\space\space3-shot\end{small} & 67.10\% & 77.00\% & 0.717 & 0.792 \\
    \begin{small}\space\space5-shot\end{small} & 67.40\% & 73.00\% & 0.7 & 0.777 \\
\midrule
    Justification &  &  &  &  \\
    \begin{small}\space\space0-shot\end{small}  & 69.11\% & 75.81\% & 0.723 & \textbf{0.795} \\
    \begin{small}\space\space1-shot\end{small} & 66.40\% & 75.20\% & 0.705 & 0.782 \\
    \begin{small}\space\space3-shot\end{small} & 61.50\% & 83.00\% & 0.706 & 0.786 \\
    \begin{small}\space\space5-shot\end{small} & 58.50\% & \textbf{86.10}\% &  0.695 & 0.778 \\
\midrule
    SBS &  &  &  &  \\
    \begin{small}\space\space0-shot\end{small} & 66.00\% & 64.00\% & 0.653 & 0.736  \\
    \begin{small}\space\space1-shot\end{small} & 67.00\% & 76.00\% & 0.713 & 0.788 \\
    \begin{small}\space\space3-shot\end{small} & 63.20\% & 79.20\% & 0.703 & 0.782 \\
    \begin{small}\space\space5-shot\end{small} & 60.80\% & 81.00\% & 0.694 & 0.776 \\
\bottomrule
\end{tabular}
\vspace{-0.5em}
\caption{Performance metrics of GPT under different settings. Best performances in bold. 
% \yrl{@AD: Do you keep the confusion matrix (numbers of TP,FP,TN,FN) for DL+ROBERTa and GPT's simple-0 and Just-0. I want to try to derive CIs to see if they are indeed no significant differences.}\ahmad{simple-0: TP: 182, TN: 434, FP: 68, FN: 66 ... Just-0: TP: 188, TN: 417, FP: 85, FN: 60 ... RoBERTa: TP: 172, TN: 431, FP: 86, FN: 61}
}
\label{table4:GPTresults}
\end{table}

Furthermore, our findings reveal that additional context, presented in few-shot setting, has a negligible effect on F1 and tends to sacrifice Precision in favor of Recall. This is consistent with previous works \cite{zhao2021calibrate, beyond_fewshot}. For instance, \citet{beyond_fewshot} argued that in-context examples help in task identification rather than learning, particularly in reasoning-intensive tasks. This finding potentially interprets SBS results, where additional examples clarify detailed instructions and contribute to performance enhancement.

% few-shot improved results: general-purpose recommendation systems \citet{fewshot_recommendation}, extracting clinical information \citet{agrawal2022large}

% \yrl{@AD: The following paragraph is not informative. I will revise this paragraph after you modify the examples in Table 6. See my comments in Table 6.}
The results of the Justification settings offer a pathway for further investigating whether GPT's labels were produced with proper rationales. To see this, we conducted a theme analysis on GPT's results under the Justification-zero-shot setting (which yielded the best AUC).  

We grouped all the annotated samples into true/false positive/negative (TP/FP/TN/FN) according to GPT's predicted labels, and then manually reviewed all generated justifications. We summarize the five most common patterns in GPT's justifications with illustrative examples (labeled as S1--S8). Note that the prediction of our best classifier are consistent with the ground-truth labels in all of the following examples. 
% \yrl{@AD: I'm not sure what's the best way to save space, but please help fill in the example and reasoning text in the following.} \ahmad{Filled missing content. Removed Table6}

\paragraph{Correct reasoning} GPT performs well in detecting conspiratorial text relating to well-known conspiracy theories such as COVID-19, UFOs, the New World Order, elites, climate change, and CTs seen in the political domain. The justifications demonstrated its ability to connect existing resources to recognize CTs in a given post. In some instances, it is able to identify less common CTs (S1) and distinguish benign CT-related text (S2).
\begin{itemize}
\item[{\bf S1}] \tpbox{TP} (Label: Yes; GPT: Yes) \stext{Dr. Strangelove was right, except with chlorine instead of fluoride. Semen is not supposed to smell like bleach, chlorine and chlorides are getting absorbed when you bathe in/drink/inhale it. So your `precious bodily fluids' is indeed being attacked.} \rtext{The text suggests that there is a deliberate attack on people’s bodily fluids through the absorption of chlorine and chlorides when bathing, drinking, and inhaling. It also references Dr. Strangelove, a fictional character known for his conspiracy theories in the movie of the same name. The claim that semen is not supposed to smell like bleach adds to the conspiratorial nature of the text.}
\item[{\bf S2}] \tnbox{TN} (Label: No; GPT: No) \stext{Does Epstein have any links to a government intelligence agency? I’ve seen a lot of talk about this but not any actual links.} \rtext{The text is a question seeking information about whether there are any known links between Jeffrey Epstein and a government intelligence agency. It does not make any specific claims or promote a conspiracy theory; instead, it appears to be an inquiry seeking clarification or evidence regarding a topic of interest.}
\end{itemize}

\paragraph{Fail: Criticism} In some instances, GPT overinterpreted the text as conspiratorial based on keywords such as ``media'' (S3) that may have appeared in other CTs. We observed, however, that these cases were not intended to promote any CTs, but rather to provoke ideas{\rebuttal{, share opinions, or provide }} criticisms in a different context. 
\begin{itemize}
\item[{\bf S3}] \fpbox{FP} (Label: No; GPT: Yes) \stext{The corruption in science and the censorship in the media is one and the same: peer review. The notion that ideas derive their truth from the opinion of a minority able to control the public discourse, rather the idea's factuality, or merit. That's why science has been unable to produce major breakthroughs for over a century. Original thought never comes through consensus, but from fringe minds.} \rtext{The text suggests that there is corruption in science and censorship in the media, and attributes this to peer review. It claims that a minority is able to control public discourse and that science has been unable to produce major breakthroughs for over a century because original thought does not come from consensus but from fringe minds. These claims lack substantial evidence and rely on the assumption that there is a deliberate effort to suppress original ideas and control the scientific and media landscape. This aligns with the characteristics of a conspiracy theory, which typically involves the belief in secret plots or hidden agendas by powerful groups.}
\end{itemize}

\paragraph{Fail: Sarcasm} We also discovered that GPT was incapable of recognizing sarcasm and falsely interpreted the author of a post as believing in a situation that the author intended to ridicule (S4).
\begin{itemize}
\item[{\bf S4}] \fpbox{FP} (Label: No; GPT: Yes) \stext{I think we should just humor Q and MAGA and agree with them that DJT is still president and still calling the shots via FEMA etc. What harm could it do? Next election when Kamala and AOC are elected we will tell the the same thing that DJT is actually the real president. Why not?} \rtext{The author believes that Donald Trump is still in charge and making decisions through FEMA (Federal Emergency Management Agency) despite Joe Biden being the current president. This belief goes against the accepted reality of the current political situation and presents an alternative narrative without any substantial evidence. The idea that Trump is secretly controlling the government and will continue to do so even after the next election is not supported by factual information, making it a conspiracy theory.}
\end{itemize}

\paragraph{Fail: Non-existent connection (causal hallucination)} In cases where the text is limited (i.e., a few words or a single sentence), GPT appears to relate the content subjects to other distant materials, such as suspecting an action or event as conspiratorial based on the position of a mentioned person (S5) or source (S6), despite the fact that no such connection exists in the text, and no evidence for the authors' intent to support any conspiratorial beliefs.
\begin{itemize}
\item[{\bf S5}] \fpbox{FP} (Label: No; GPT: Yes) \stext{Michael Flynn Calls for Trump to Declare Martial Law.} \rtext{The statement suggests that Michael Flynn is calling for Trump to declare martial law, which implies a potential conspiracy theory. The idea of a former high-ranking military officer advocating for martial law can be seen as a controversial and potentially conspiratorial claim. However, without further context or evidence, it is difficult to determine the veracity of the statement.}
\item[{\bf S6}] \fpbox{FP} (Label: No; GPT: Yes) \stext{Reporter Who Broke Clinton-Lynch Tarmac Story And Received Significant Death Threats Found Dead. See Infowars (not linking to it in Reddit).} \rtext{The statement suggests that the reporter who broke the Clinton-Lynch Tarmac story and received death threats has been found dead. However, without any further evidence or verifiable sources, it is difficult to determine the truthfulness of this claim. The mention of Infowars also raises suspicions, as Infowars is known for promoting conspiracy theories. Therefore, until there is more credible evidence to support this claim, it can be considered a conspiracy theory.}
\end{itemize}

\paragraph{Fail: Implicit connection or rhetorical question} In contrast to its tendency to overinterpret or misuse subjects that appear in the text (e.g., keywords, well-known officials, and media outlets), GPT is incapable of recognizing more nuanced communication cues, such as an implicit attribution {\rebuttal{(S7)}} or a rhetorical question {\rebuttal{(S8)}}, which are frequently observed in online texts. {\rebuttal{In the next two examples, we provide human reason due to the subtlety of these cues.}} 
\begin{itemize}
\item[{\bf S7}] \fnbox{FN} (Label: Yes; GPT: No) \stext{Turkey rocked by 7M Quake hours after France attack; World leaders Outraged at Macron!} \rtext{The given text does not provide any evidence or support for a conspiracy theory. It mentions two separate events, a 7M earthquake in Turkey and an attack in France, along with world leaders being outraged at Macron. While the text may imply a connection between these events, it does not present any conspiracy theories or elaborate on any hidden agendas or secret plots. Therefore, it cannot be considered a conspiracy theory.} {\rebuttal{\underline{Human reason}: ``The author connects two unrelated events, which indicates that the first is caused by the second. This argument is supported by the last sentence of the post, which states that world leaders are outraged at France accusing them of committing the earthquake.''}}
\item[{\bf S8}] \fnbox{FN} (Label: Yes; GPT: No) \stext{How does marxism/communism take over so many countries? It happened to Europe, the UK, Russia and is now happening to the US. How are they able to do this to every country while so many people don't even realize it?} \rtext{The statement does not present any conspiracy theories. It discusses the spread of Marxism/Communism to various countries and questions how it has been able to happen without many people realizing it. While it may touch upon political ideologies and their influence, it does not propose any hidden or secretive plots, making it not a conspiracy theory.} \newline
{\rebuttal{\underline{Human reason}: ``The author addresses the spread of Marxism/communism in a systematic method. The question is rhetoric and implies an affirmative tone. This aligns with the known CT `Cultural Marxism', which talks about the spread of Marxism in Europe, and extends it to the US.''}}
\end{itemize}

Based on our analysis of GPT's classification reasoning, it was found that, in certain cases, GPT can accurately identify the narrative elements that support its decision to classify a text as either CT or non-CT. However, there are instances where it misuses information both within and outside the text, and fails to recognize the subtle dialectic cues often present in informal communications. While GPT has been shown useful in other domains, such as data augmentation \cite{moller2023prompt}, our study suggests that it should be used with caution in the context of CTs. 
% \yrl{@AD: to add citation} \ahmad{Done}

% Furthermore, our most proficient classifier, i.e. RoBERTa in Sec.\ref{sec7.1}, achieved comparable F1 results across all GPT prompt settings and outperformed them in Precision, except for Simple-Zeroshot. These outcomes underscore the potential limitations of GPT as an approach to labeling online conspiratorial content. Consider Table \ref{table6:GPTlabels}, GPT exhibits strengths in its ability to identify less common CTs (Example1), and discern benign text referencing CT-related topics (Example2). However, it faces challenges in recognizing newly proposed CTs (Example3) and narratives concealed within rhetorical questions (Example4), categorizing the former as hypothetical scenarios and the latter as legitimate discussion topics. Additionally, GPT displays a propensity for labeling texts expressing opinions or frustrations as conspiratorial (Example5), and it grapples with detecting sarcasm (Example6). Notably, our classifier correctly labeled all the aforementioned examples, underscoring the potential necessity of a supervised approach to capture the nuanced language and contextual intricacies found in online conspiracy discourse. This conclusion motivates our decision to further explore this model in subsequent experiments in the next section.

\subsection{Prevelance of CT Narratives in CT-Subreddits}
\yrv{We investigate the prevalence of CT narratives within the most active online conspiracy subreddits as described in Sec.~\ref{sec:data}. We employ the same filtering procedures outlined in Sec.\ref{sec5.1}, which include removing self-deleted, banned, and short posts. We then use our best-trained classifier, i.e. RoBERTa, to classify each of the posts into CT or non-CT. Table \ref{table5:full_classification} lists the positive ratios, i.e., proportions of conspiratorial narratives, for each of the 12 subreddits. We estimate the upper and lower bounds of the positive ratios based on the Precision and Recall of the classifier, as well as the detected positive ratios and sample size. Specifically, the upper bound is estimated by assuming that all detective positives are true positives, and the lower bound is estimated by assuming the detective positives contain no false negatives. We observe a wide range of positive ratios, from 20\% to 46.5\%. It is important to note that these ratios may be affected by the distinct focus of each subreddit as well as the moderation in place. The top three subreddits with the highest positive ratios are r/1984isreality (46.5\%), r/conspiracyundone (41.9\%), and r/TopConspiracy (40.5\%). The ratio of all subreddit posts, 31.3\%, corresponds closely to the ratio observed in the annotated subset, 33\%.}
% To investigate the prevalence of narratives within online conspiracy subreddits, we leveraged the fully assembled dataset described in Sec.\ref{sec:data}. Following the same filtering procedures outlined in Sec.\ref{sec5.1}, which involved removing deleted, banned, and short posts, we employed our best-trained classifier, i.e. RoBERTa, to classify all posts into CT and non-CT. The findings, as presented in Table \ref{table5:full_classification}, reveal that the proportion of conspiratorial content varies across all subreddits, ranging from 20\% to 46.5\%. It is worth noting that these percentages may be influenced by the unique focus of each subreddit and the moderation in place. Interestingly, the overall ratio of the predicted conspiratorial content across all subreddits, at 31.3\%, closely aligns with the ratio observed within the annotated subset, 33\%.

% To further assess the robustness of our results, we conducted an estimation of the upper and lower bounds for the positive sample ratio. This estimation relied on both the Precision and Recall of our best classifier, in conjunction with the positive sample ratio of each subreddit. 
% % \yrl{@AD: That's the nature of the math. Nothing intriguing.}\ahmad{Removed the sentence}

% \yrv{In contrast with prior works ...}\yrl{@AD or NR: Plz elaborate how the ratio (overall) is different compared with prior works.}
%We used the following equations:
%\begin{center}
%$UpperBound_i = Pr * Rp_i / Re $
%\label{eq:upper_bound}
%\end{center}
%\begin{center}
%$LowerBound_i = Pr * Rp+i $
%\label{eq:lower_bound}
%\end{center}

\begin{table}[h]
\small
\centering
\begin{tabular}{l r m{2em} m{2em} m{2em}}% l l}
\toprule
    \textbf{Subreddit} & \textbf{Posts} & \textbf{Pos. \newline Ratio} & \textbf{Upper Bound} & \textbf{Lower Bound} \\ %& \textbf{upper neg bound} & \textbf{lower neg bound}\\
\midrule
conspiracy & 201054 & 0.312 & 0.422 & 0.218 \\%& 0.811 & 0.745\\
TruthLeaks & 872 & 0.279 & 0.377 & 0.195 \\%& 0.853 & 0.801 \\
TopConspiracy & 311 & 0.405 & 0.547 & 0.284 \\%& 0.755 & 0.669 \\
conspiracy\_commons & 12941 & 0.321 & 0.434 & 0.225 \\%& 0.818 & 0.754 \\
climateskeptics & 3078 & 0.235 & 0.318 & 0.165 \\%& 0.853 & 0.801 \\
conspiracytheories & 11379 & 0.337 & 0.455 & 0.236 \\%& 0.825 & 0.764 \\
DescentIntoTyranny & 121 & 0.273 & 0.369 & 0.191 \\%& 0.825 & 0.764 \\
ConspiracyII & 1059 & 0.355 & 0.480 & 0.249 \\%& 0.832 & 0.773 \\
FringeTheory & 385 & 0.200 & 0.270 & 0.140 \\%& 0.895 & 0.858 \\
conspiracyundone & 1420 & 0.419 & 0.566 & 0.293 \\%& 0.755 & 0.669 \\
C\_S\_T & 4974 & 0.318 & 0.430 & 0.223 \\%& 0.804 & 0.735 \\
1984isreality & 43 & 0.465 & 0.628 & 0.326 \\%& 0.762 & 0.678 \\
\midrule
Overall & 237637 & 0.313 & 0.423 & 0.219 \\%& 0.811 & 0.745 \\
\bottomrule
\end{tabular}
\vspace{-0.5em}
\caption{Classification results of posts in each subreddit. The second column shows the number of posts (CT and non-CT) considered in this study after filtering. 
% \yrl{@AD: Is the first column total positives, or total posts? plz clarify.} 
% \ahmad{Total, +ve and -ve. Added clarification in the caption}
}
\label{table5:full_classification}
\end{table}

% To investigate potential disparities in the comments and karma scores (the difference between upvotes and downvotes, with zero as the minimum) received by CT posts compared to non-CT posts, we conduct a comparative analysis of the distributions within this sub-corpora. Given our observation that both distributions show scale-free characteristics, we employ the empirical Cumulative Distribution Function (eCDF) for this comparison (see Figure \ref{fig1}). Our findings indicate that CT posts typically accumulate more comments and receive greater karma scores, signifying greater engagement and a higher ratio of upvotes to downvotes. These differences are significant for both the comments (Mann-Whitney U test: $6.17 \times {10^{9}}$ with a $p-value < 10^{-6}$) and karma scores (Mann-Whitney U test: $6.36 \times {10^{9}}$ with a $p-value < 10^{-6}$). This raises concerns, particularly when conspiratorial narratives are employed to disseminate animosity towards specific groups or false information, as the power of persuasion can make them even more dangerous than hate narratives or fake news alone.

% \yrl{@NR: plz elaborate why the results are significant. What are the consequence when there are significantly higher reactions to CT vs. non-CT?}

% \ana{I just added one sentence above. I'm not sure if that argument good enough though.}

% \ana{ @YRL: BTW, should I report the two-sided or one-sided Mann-Whitney scores in this case? I have the results of both.}

\begin{figure}[t]
\centering
\includegraphics[width=1\columnwidth]{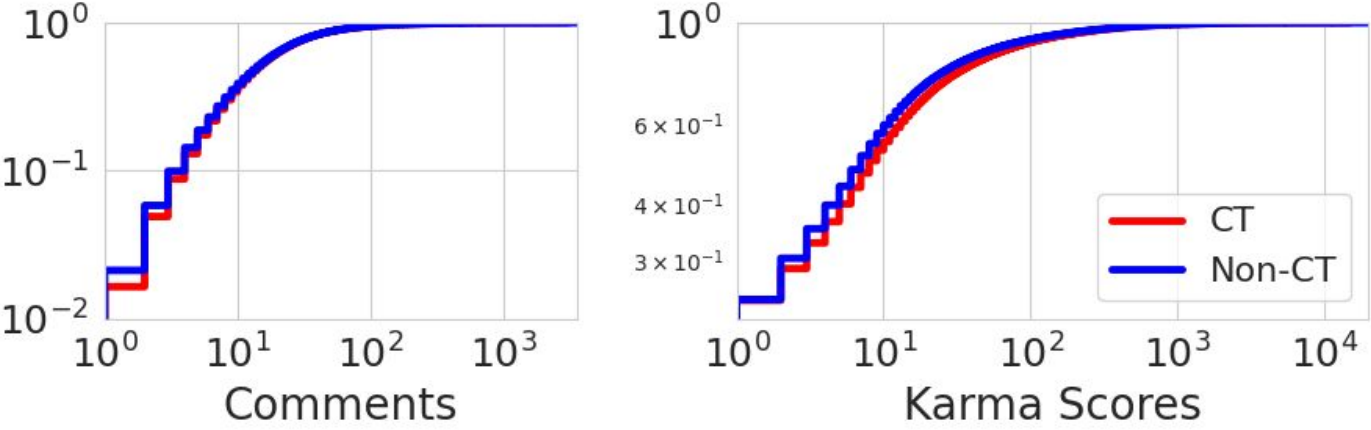}
\vspace{-1em}
\caption{Disparities of comments (left) and karma scores (right) distributions between CT and non-CT posts, represented by the eCDF.}
\label{fig1}
\vspace{-1em}
\end{figure}

We further analyze the differences of CT and non-CT posts in terms of audience reactions they received. We use two measures for the reactions: (a) number of comments, and (b) karma scores, defined as the number of upvotes minus the number of downvotes, or zero if the difference is negative. We compare the distributions of the two measures in our dataset. Fig. \ref{fig1} shows the empirical cumulative distribution function, or eCDF, for the two measures. As both measures have skewed distributions, the Mann-Whitney U test is used to determine whether the two distributions are significantly different. Overall, we found that CT posts tend to receive more comments than non-CT ($p<10^{-6}$) and greater karma scores ($p<10^{-6}$).

%Despite the similar appearance of the eCDF plots of CT and non-CT posts, the test result indicates that non-CT posts typically receive more comments ($p<10^{-3}$) and higher karma scores ($p<10^{-3}$).
% Because the distributions are featured by their heavy tails, we specifically examine differences in the tails of the two distributions. We compare on the top 5\% of values for each measure. The top 5\% of comment counts vary between 102 and 3068 in CT posts and between 100 and 3593 in non-CT posts, whereas the top 5\% of karma scores vary between 209 and 14797 in CT and 180 and 19734 in non-CT posts. The result shows that the top 5\% non-CT posts also have more comments ($p<10^{-3}$) and higher karma scores ($p<10^{-6}$) than the top 5\% CT posts. \yrl{@NR fill the details, and revise the sentence if my assumption is incorrect.} \ana{I just revised my interpretation: it seems that indeed non-CT posts having more engagement}
These findings indicate that a CT post typically receives greater engagement and a higher ratio of upvotes to downvotes. On the one hand, this reflects the nature of these subreddits, where users tend to be more interested in CTs; on the other hand, it raises concerns that posts with CTs, with more interactions received, are likely to be promoted by platform algorithms, and users who post conspiratorial narratives may accumulate karma scores to spread their content more easily than those who do not (e.g., actively debunk CTs).

%The results are interesting. On the one hand, we anticipated that users of these subreddits would be more interested in posts promoting CTs, but the results show that non-CT posts did not receive less attention. On the other hand, the data we studied may have been moderated. In the remaining posts, conspiratorial narratives with fewer interactions are less likely to be promoted by platform algorithms, and users who post conspiratorial narratives may not accumulate karma scores to spread their content more easily than those who do not (e.g., those who actively debunk conspiratorial narratives). }

\subsection{Cross-domain Test}
{\rebuttal{To assess the efficacy of our method on samples from different domains, we employ our top-performing model, RoBERTa, on a Twitter dataset obtained from \cite{phillips2022hoaxes} and conduct a comparative analysis. In their study, a two-step approach was adopted: the first step detects the presence of a CT in a tweet, while the second step classifies the tweet into one of three categories—Against, Neutral, and Supportive. In our work, we consolidated the first step and the first two categories into non-CT, making the Supportive classification directly comparable to CT content only. The results are presented in Table \ref{table6:generalizability}.

It is worth noting that our classifier achieved a high F1 score despite not receiving any training or fine-tuning on this Twitter dataset. In fact, the score was only slightly lower than the authors' best model which was specifically trained using the same dataset. Furthermore, our classifier demonstrated an impressive precision rate of 91\%. However, it was expected to have a lower recall rate, given that the posting styles on the two platforms differ significantly. This suggests that our approach, which integrates attitudes towards a conspiracy theory into a unified classifier, makes it possible to identify conspiratorial posts more precisely, even across different social media platforms and topics.}}

\begin{table}[h]
\small
\centering
\begin{tabular}{l l m{2em} m{2em} m{2em}}
\toprule
    \textbf{Model} & \textbf{Precision} & \textbf{Recall} & \textbf{F1} \\ 
\midrule
\citet{phillips2022hoaxes} & 71.1\% & 78.5\% & 0.746 \\
Our RoBERTa & 91\% & 56\% & 0.690 \\
%\midrule
%Overall & 237637 & 0.313 & 0.423 & 0.219 \\
\bottomrule
\end{tabular}
\vspace{-0.5em}
\caption{Cross-Domain Comparison of CT Classifications}
\label{table6:generalizability}
\end{table}

\section{Discussions} \label{sec8}
This research examined the pervasive problem of detecting conspiracy theories in online discussions. We developed a comprehensive and general classification scheme that incorporates the theoretical and operational definitions of conspiracy theories, as well as deep learning and large language models, to distinguish conspiratorial narratives effectively. Our approach utilized human-labeled ground truth to train a BERT-based classifier. This classifier was subsequently used to examine the ratios of conspiratorial narratives in the most active conspiracy-related Reddit forums. Our research revealed that only one-third of these forums' posts were classified as containing conspiracy theory narratives. This finding challenged previously held assumptions regarding the prevalence of conspiratorial narratives in such communities {\revisedrebuttal{\cite{phadke2021characterizing, gram_ct,klein2019pathways,engel2022characterizing, papasavva2021qoincidence, bessi2015science,zollo2017debunking}}}.

Our research also revealed that posts that promote CT narratives tend to receive more comments and higher karma scores, indicating that CT promoters may gain additional advantages that enable them to promote their content more broadly. Based on our findings, platforms should consider a different promotion mechanism that takes into account the distinct nature of the online communities, and more sophisticated techniques should be integrated into content moderation to reduce the visibility of conspiratorial content.

% Our research also revealed that posts that promote non-CT narratives tend to receive more comments and higher karma scores, suggesting that CT promoters would face difficulties in promoting their content more broadly, assuming proper moderation was in place. Therefore, more sophisticated techniques should be integrated into content moderation in order to further reduce the visibility of conspiratorial content.

%the fact that GPT was 
In addition, our analysis showed that our classifier performs comparably to GPT, despite being trained on extensive text corpora and computational resources. Our qualitative analysis of GPT's classification decision justifications revealed several alarming flaws, indicating that its capacity for enhancing data quality or detection performance in the context of CT is limited. %\ana{do we need to remove this part (above) now because the non-CT posts are the ones having higher engagement? :( } 

This study has some limitations. Firstly, the classification criteria aimed to detect both established CTs and content with the potential to evolve into such theories. However, the inherent ambiguity of authors' intentions in online discussions, particularly in relation to conspiracy beliefs, made accurate categorization challenging. Second, despite our efforts to ensure consistency in annotation, different annotators may interpret specific samples differently, potentially introducing subjectivity into the annotation process. Lastly, while the extent of GPT's uncontrollable randomness that drives the diverse outcomes may not significantly affect the average performance evaluation, it may pose a reproducibility challenge. Future research should consider methods that account for the variability of GPT's responses in order to systematically improve its ability to perform tasks requiring a substantial amount of contextual understanding.

\section{Ethical Considerations}\label{sec:ethics}

\yrv{\textbf{Data Collection and Management.} The data used in this study were collected through the Pushshift Reddit API. We used the Pushshift API to access publicly available subreddits and posts that users chose to make public on Reddit. We also collected publicly viewable metrics related to publicly posted content (e.g., a post's ''comment'' count and ''score'' value). All of the information we collected has been shared publicly on the platform with an unrestricted audience. Starting June 19, 2023, access to data via third-party services was limited per Reddit's introduction of new Data API Terms.\footnote{\url{https://www.redditinc.com/policies/data-api-terms}} As of the time this paper was written, Pushshift only provided restricted access.\footnote{\url{https://www.reddit.com/r/pushshift/comments/13w6j20/advancing_communityled_moderation_an_update_on/}} We adhere to Reddit's new policy and the platform's data usage guidelines,\footnote{\url{https://support.reddithelp.com/hc/en-us/articles/14945211791892#h_01H69EJB9GRHCMPZMKFQTNQKY0}} which state that the data may only be used for research purposes, and we will not redistribute our data or any derivative products or services based on our data (e.g. models trained using Reddit data) without further permission from Reddit. Prior to publishing our work, we will seek Reddit's permission for data usage and possible academic sharing.
}

\yrv{\textbf{User Anonymity and Privacy.} 
We had no direct interaction with Reddit users and gathered no private information about them. As outlined in Sec.~\ref{sec:data}, we excluded the self-deleted or banned posts in our analysis. Consequently, our study may be biased and may have missed some of the most harmful narratives. Our study results were either presented anonymously or in a summary format, with no user-specific information disclosed. While we included a few Reddit posts as examples, we took measures to ensure the anonymity of these sources. Specifically, we conducted a comprehensive search (using the keyword ``Reddit'' as well as the post's title and content on a popular search engine) to ensure that the user information associated with the presented examples cannot be easily recovered.}

\yrv{\textbf{Content Credibility}. Several examples of conspiracy narratives are presented in this paper for illustrative purposes, which poses a risk that some readers may consider them credible. Even though some CTs turn out to be true, we emphasize that a vast number of CTs are not credible  \cite{ct_df1} -- they are unproven, misleading, or lack empirical evidence. We caution our readers to read these examples with skepticism and critical thinking.}%In light of this, we caution our readers to read these examples with skepticism and critical thinking and to evaluate their validity independently in any given context.}

\yrv{\textbf{Annotation Complexity and Subjectivity.} {\revisedrebuttal{\iffalse Annotating online CT content, although valuable, presents inherent complexities. The absence of clear authorial intentions and the challenge of predicting whether a proposed narrative may evolve into a CT or remain benign contribute to the difficulty. Additionally, the inherent subjectivity in interpreting specific texts poses a challenge to precise labeling. Despite our endeavors to achieve consensus, we acknowledge that other researchers may arrive at different decisions. \fi
In accordance with our proposed coding scheme, a comprehensive, topic-independent annotation dataset has been generated, which can be utilized by researchers in the future. However, we acknowledge that the annotation process is inherently subjective, and human errors may arise due to its complexity. For instance, determining the attitude of a post-author can be challenging, particularly when authors deliberately conceal their attitudes. Despite striving to reach a consensus among our well-trained annotators, it is acknowledged that other researchers may arrive at different decisions.}}}

\yrv{\textbf{Potential Use/Abuse of Work.} This study aimed to understand the feasibility and limitations of identifying online conspiracy narratives. While our work contributed to ways of countering the spread of conspiracy theories and promoting a more healthy public discourse, there is a possibility that individuals who intend to disseminate CTs may take advantage of our study outcome to bypass automatic detection mechanisms. By acknowledging this limitation, we underscore the importance of ongoing research and vigilance in refining automated detection methods to safeguard against potential abuse of new research outcomes.}

% \section*{Acknowledgement}
\noindent\paragraph{Acknowledgement}
The authors would like to acknowledge support from AFOSR, ONR, Minerva, NSF \#2318461, Collaboratory Against Hate Research and Action Center, and Pitt Cyber Institute's PCAG awards. The research was also partially supported by Pitt's CRC resources (RRID:SCR 022735 through NIH \#S10OD028483).
Any opinions, findings, and conclusions or recommendations expressed in this material do not necessarily reflect the views of the funding sources.

\bibliography{reference}

\newpage

\end{document}